\documentclass[14pt,letterpaper]{article}  
\usepackage[cp1251]{inputenc} 
\usepackage[T2A]{fontenc}     
\usepackage[russian,english]{babel} 
\usepackage{graphicx}
\usepackage{float}
\usepackage{wrapfig}
\usepackage{longtable}

\usepackage[tableposition=top]{caption}

\usepackage{tabularx}
\usepackage{array,multirow}
\newcommand{\tabularxmulticolumncentered}[3] 
    {\multicolumn{#1}
                 {>{\centering\hsize=\dimexpr#1\hsize+#1\tabcolsep+\arrayrulewidth\relax}#2}
                 {#3}}

\usepackage{extsizes}
\usepackage{geometry}
\title{\textbf{A big data approach towards sarcasm detection in Russian}}  
\date{}  
\author{Gurin A.A., Sadykov.T.M., Zhukov T.A.}

\begin{document}
\bibliographystyle{splncs04}
\renewcommand\refname{\centering References}  

\newcommand{\CS}{C\nolinebreak\hspace{-.05em}\#}

\hyphenation{Open-Cor-pora}
\maketitle  
  
\begin{center}Plekhanov Russian University of Economics\end{center}
 
\textbf{Abstract:}
 \newline
\par We present a~set of deterministic algorithms for Russian inflection and automated text synthesis.
These algorithms are implemented in a~publicly available web-service www.passare.ru.
This service provides functions for inflection of single words, word matching and synthesis of grammatically correct Russian text. Selected code and datasets are available at  https://github.com/passare-ru/PassareFunctions/
\par Performance of the inflectional functions has been tested against the annotated corpus of Russian language \texttt{OpenCorpora}~\cite{openCorpora}, compared with that of other solutions, and used for estimating the morphological variability and complexity of different parts of speech in Russian.
 \par \textbf{keywords:} Natural language generation, automatic text synthesis,algorithmic inflection of Russian, morphological variability, Sarcasm detection.
 \newline

\par\textbf{\large Introduction}
\par Automatic inflection of words in a~natural language is necessary for a~variety of theoretical and applied purposes like parsing, topic-to-question generation~\cite{Chali-Hasan}, speech recognition and synthesis, machine translation~\cite{IomdinOliverSagalova}, tagset design~\cite{Kuzmenko}, information retrieval~\cite{Iomdin}, content analysis~\cite{Belonogov,BelonogovKotov}, and natural language generation~\cite{CeruttiEtAl,CostaEtAl,RajeswarEtAL,Tran}. Various approaches towards automated inflection have been used to deal with particular aspects of inflection~\cite{EnglishPlural,Zaliznyak} in predefined languages~\cite{Foust-AutomaticEnglishInflection,Latin,Sanskrit,RussianAndUkrainian,Porter} or in an unspecified inflected language~\cite{languageFree,Silberztein-NooJ}. 
\par Despite substantial recent progress in the field~\cite{Ascoli,Buddana2021,RussianAndUkrainian,Silberztein-NooJ,Sorokin,Xiao-Zhu-Liu}, automatic inflection and automatic text generation still represent a~problem of formidable computational complexity for many natural languages in the world. Most state-of-the-art approaches make use of extensive manually annotated corpora that currently exist for all major languages~\cite{Segalovich}. Real-time handling of a~dictionary that contains millions of inflected word forms and tens of millions of relations between them is not an easy task~\cite{Goldsmith}. Besides, no dictionary can ever be complete. For these reasons, algorithmic coverage of the grammar of a~natural language is important provided that inflection in this language is complex enough.
\par
\begin{figure*}[ht!]
\centering{}\includegraphics[width=\textwidth]{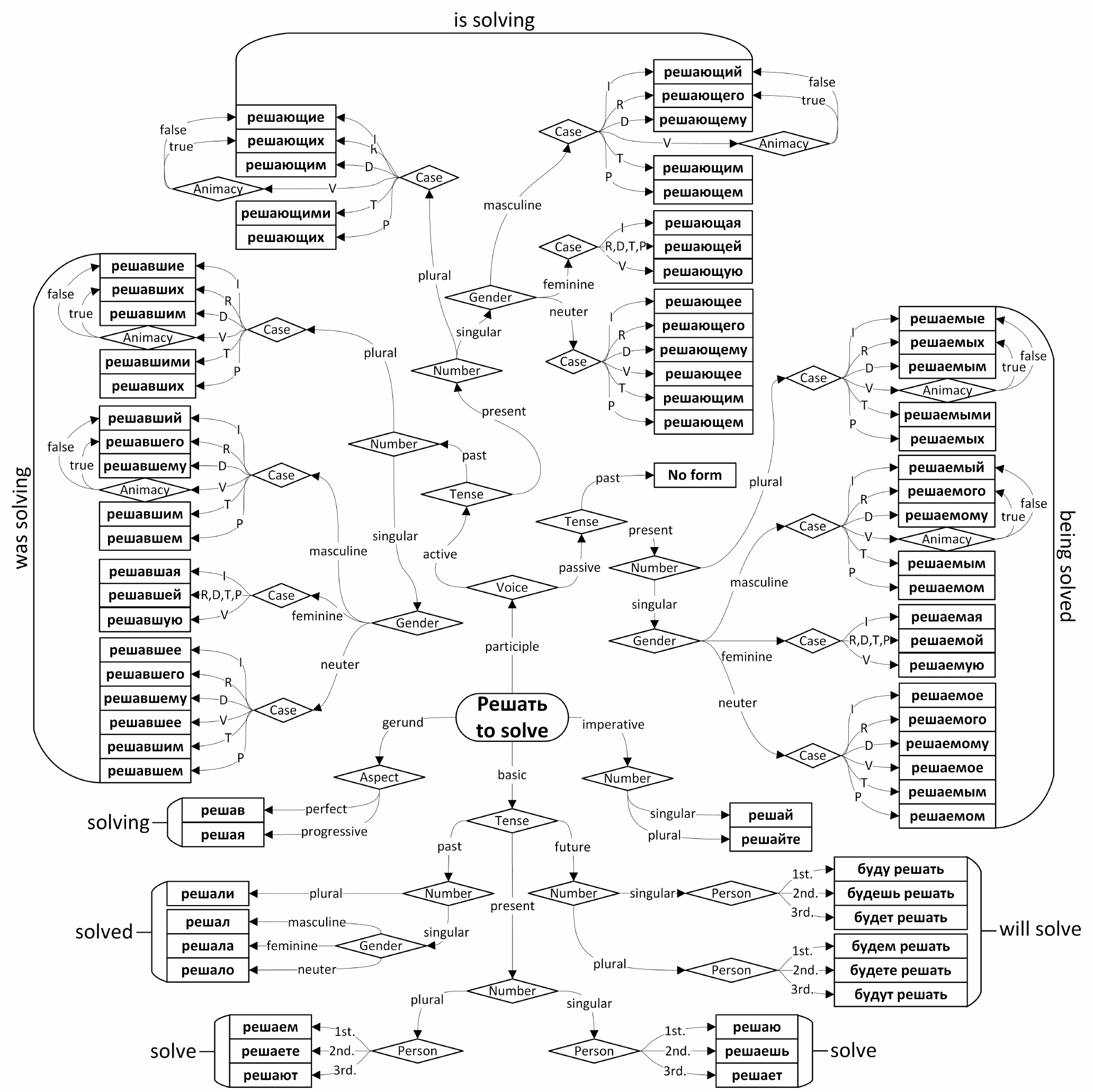}
\caption{All of the forms of the Russian verb
"\textcyrillic{решать}" \textemdash  "reshat${}^{'}\,$" \textemdash  "to solve"} and their dependence on tense (past, present, or future), number (singular or plural), gender (feminine, masculine, or neuter), person (1st, 2nd, or 3rd), voice (active or passive), aspect (perfect or progressive), case (nominative, genitive, dative, accusative, instrumental, or prepositional,
abbreviated in accordance with Russian translation by I, R, D, V, T, and P, respectively), and animacy (boolean-valued)
\label{fig:Reshat-AllForms}
\end{figure*}

 Russian is a~highly inflected language whose grammar is known for its complexity~\cite{Sorokin,Zaliznyak}. In~Russian, inflection of a word may require changing its prefix, root, and ending simultaneously while the rules of inflection are highly complex~\cite{Halle-Matushansky,Zaliznyak}. The form of a~word can depend on as~many as~seven grammatical categories such as~number, gender, person, tense, case, voice, animacy etc (cf Fig.~\ref{fig:Reshat-AllForms}). By~an estimate based on~\cite{openCorpora}, the average number of different grammatical forms of a~Russian adjective is~11.716. A~Russian verb has, on average,~44.069 different inflected forms, counting
participles of all kinds and the gerunds (cf.~Fig.~\ref{fig:Reshat-AllForms}).
\par In the present paper we describe a~fully algorithmic dictionary-free approach towards automatic inflection of Russian. The algorithms exposed in the paper are based on morphological rules that have been manually created and tuned by the authors. Implementation of the algorithms has been performed in C\# programming language, see 
https://github.com/passare-ru/PassareFunctions/ The described functionality is freely available online at www.passare.ru through both manual entry of a~word to be inflected and by API access of main functions for dealing with big amounts of data.

\section{Inflection in Russian Language: Algorithms and Implementation}
\label{sec:implementation}

\begin{figure*}[ht!]
\centering{}
\includegraphics[width=\textwidth]{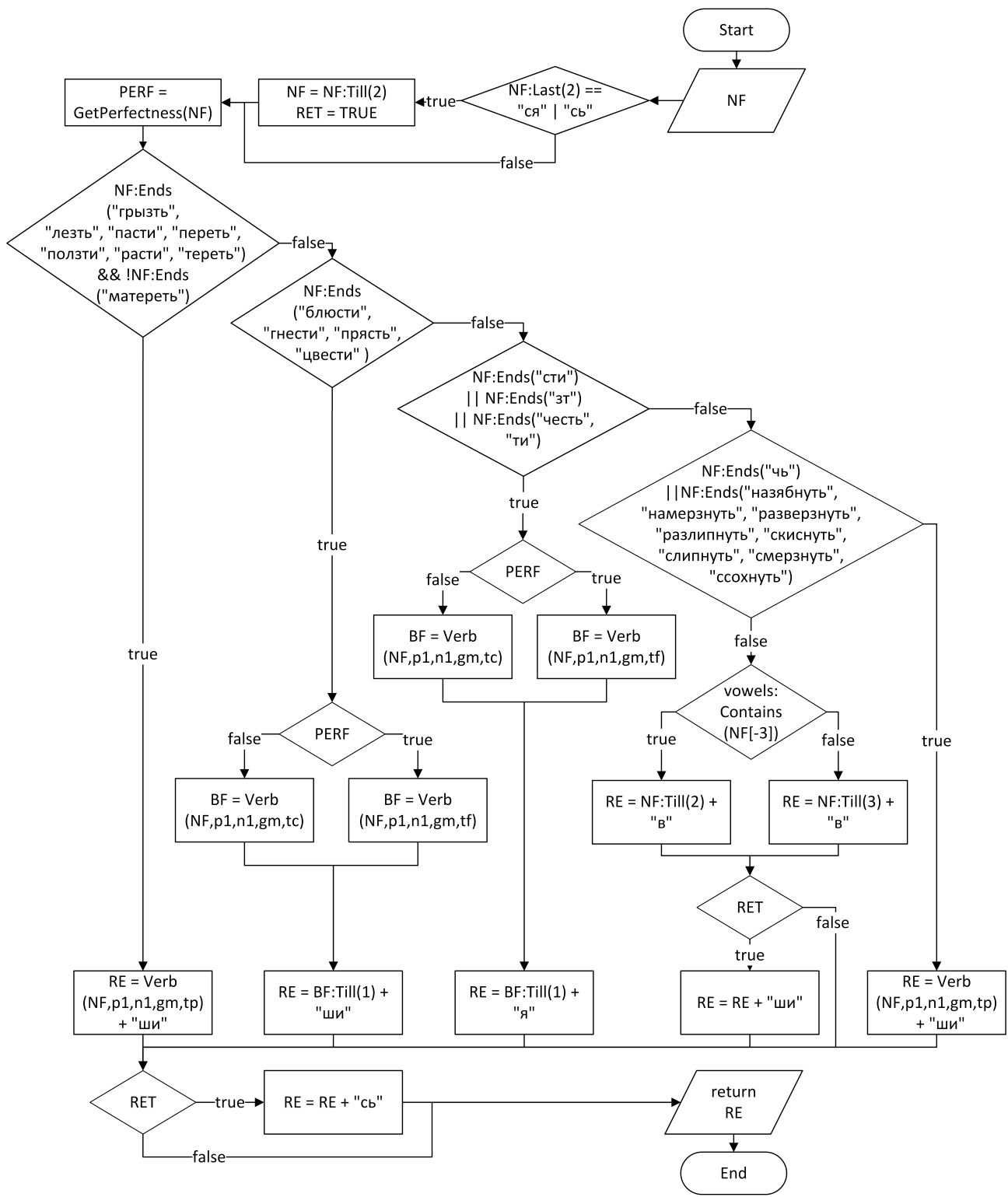}
\caption{Generation of the perfective gerund form of a~verb.} 
\label{fig:perfectiveGerundFlowchart}
\end{figure*}

Russian words and letters are given in conventional English transliteration. 
\texttt{NF, BF,} and \texttt{RE} stand for the normal form of a verb, the base of a verb, and the result of the computation, respectively. 
\texttt{PERF} and \texttt{RET} are boolean variables encoding perfective and reflexive properties of a verb, respectively. 
The list \texttt{vowels} comprises Russian vowels. The other notation coincides with the \CS\, syntax, NF:Till(2) standing for the string NF will two last characters removed.
The corresponding \CS\, code is available at https://github.com/passare-ru/PassareFunctions/

\par The web-service passare.ru offers a~variety of functions for inflection of single Russian words, word matching, and synthesis of grammatically correct text. In~particular, the inflection of a~Russian noun by number and case, the inflection of a~Russian adjective by number, gender, and case, the inflection of a~Russian adverb by the degrees of comparison are implemented. Russian verb is the part of speech whose inflection is by far the most complicated in the language. The implemented algorithms provide inflection of a~Russian verb by tense, person, number, and gender. These algorithms also allow one to form the gerunds and the imperative forms of a verb. Besides, functions for forming and inflecting active
present and past participles as~well as~passive past participles have been implemented. Passive present participle is the only verb form not currently supported by the website due to the extreme level of its irregularity. Besides, passive present participle cannot be formed at all for numerous verbs in the Russian language.
\par The algorithmic coverage of the Russian language provided by the web-service passare.ru aims to balance grammatical accuracy and ease of use. For that reason, a~few simplifying assumptions have been made: the Russian letters "\textcyrillic{\"е}" and "\textcyrillic{е}" are treated as identical; no information on the stress in a word is required to produce its inflected forms; for inflectional functions, the existence of an input word in the language is determined by the user. Furthermore, the animacy of a~noun is not treated as~a~variable category in the noun-inflecting function despite the existence of~1037 nouns (about~1.4\% of the nouns in the \texttt{OpenCorpora} database~\cite{openCorpora})
with unspecified animacy. This list of nouns has been manually reviewed by the authors on a~case-by-case basis and the decision has been made in favor of the form that is more frequent in the language than the others. The other form can be obtained by calling the same function with a~different \texttt{case} parameter (\texttt{Nominative} or \texttt{Genitive}
instead of \texttt{Accusative}).

\par Similarly, the perfectiveness of a verb has not been implemented as~a~parameter in a verb-inflecting function although by~\cite{openCorpora} there exist~1038 verbs (about 3.2\% of the verbs in the database) in the language whose perfectiveness is not specified. For such verbs, the function produces forms that correspond to both perfective and
imperfective inflections.

\begin{table*}[ht!]
\centering
\centerline{Table~1: Inflection speed and agreement rates of passare.ru and \texttt{OpenCorpora}}
\label{tab:timesOfInflection} {\small \vskip0.2cm \noindent
\begin{tabular}{|m{2cm}|m{2.5cm}|m{1.9cm}|m{1.8cm}|m{1.7cm}|m{2,3cm}|}
\hline
Part of speech        & Total number of words       & Total processing time, min:sec      & Number of forms computed (per word) & Processing time per word, msec & Agreement rate with \texttt{OpenCorpora}\\
\hline Noun           & 74633                       & 02:36                               & 12    & 2      &  98.557\,\%    \\
\hline Verb            & 32358                       & 05:49                               & 24    & 10     &  98.678\,\%    \\
\hline Adjective      & 42920                       & 00:06                               & 28    & 0.14   &  98.489\,\%    \\
\hline Adverb         & 1507                        & $<$00:01                            & 2     & 0.021  &  n/a           \\
\hline Ordinal        & 10000 \mbox{(range~0-9999)} & 00:30                               & 18    & 3      &  n/a           \\
\hline Cardinal       & 10000 \mbox{(range~0-9999)} & 00:23                               & 24    & 2      &  n/a           \\
\hline Present participle \mbox{active}  & 16946    & 04:55                               & 28    & 17     &  98.961\,\%    \\
\hline Past participle \mbox{active}       & 32358    & 10:19                               & 28    & 19     &  99.152\,\%    \\
\hline Past participle \mbox{passive}     & 32358    & 10:32                               & 28    & 19     &  94.803\,\%    \\
\hline Gerunds                                    & 32358    & 00:23                               & 2     & 0.72   &  99.157\,\%    \\
\hline Verb imperative                         & 32358    & 00:42                               & 2     & 1      &  95.327\,\%    \\
\hline
\end{tabular}
}
\end{table*}

\par The inflectional form of a~Russian word defined by a~choice of grammatical categories (such as~number, gender, person, tense, case, voice, animacy etc.) is in general not uniquely defined. This applies in particular to many feminine nouns, feminine forms of adjectives, and to numerous verbs. For such words, the algorithms implemented in the web-service passare.ru only aim at finding one of the inflectional forms, typically, the one which is the most common in the language.
\par Due to the rich morphology of the Russian language and to the high complexity of its grammar, a~detailed description of the algorithms of Russian inflection cannot be provided in a~short research paper. The algorithm for the generation of the perfective gerund form of a~verb is presented in Fig.~\ref{fig:perfectiveGerundFlowchart}. Using the verb "\textcyrillic{решать}" \textemdash  "reshat${}^{'}\,$" \textemdash  "to solve" as input, the algorithm outputs the gerund "\textcyrillic{решав}" \textemdash  "reshav" \textemdash  "having solved". Most of the notation in Fig.~\ref{fig:perfectiveGerundFlowchart} is the same as~that of the~\CS\, programming language. Furthermore, \texttt{NF} denotes the input normal form (the infinitive) of a~verb to be processed. \texttt{GetPerfectness()} is a~boolean function which detects whether a~verb is perfective or not. \texttt{Verb()} is the function which inflects a~given verb with respect to person, number, gender and tense. \texttt{BF} denotes the basic form of a~Russian verb which is most suitable for constructing the perfective gerund of that verb. We found it convenient to use one of the three different basic forms depending on the type of the input verb to be inflected. The list \texttt{vowels} comprises all vowels in the Russian alphabet.
\par Although Russian morphology is extensively covered in the literature, the algorithms of the web-service www.passare.ru are in general fully novel and very different from other existing algorithmic approaches or textbook rules. The implementation comprises about 35,000 lines of code and has been compiled into a~571~kB executable file.

\section{Software Speed Tests and Verification\\ of~Results}
\label{sec:speedTests}

\par The software being presented has been tested against one of the largest publicly available corpora of Russian, \texttt{OpenCorpora}~\cite{openCorpora}. We have been using Intel Core i5-2320 processor clocked at 3.00GHz with 16GB RAM under Windows~10.  With all indefinite forms of the words in the \texttt{OpenCorpora} database as input, the whole output produced by www.passare.ru has been checked against the corresponding forms in the database to see how many discrepancies are present. The results are summarized in Table~1.
\par All of the words whose inflected forms did not show full agreement with the \texttt{OpenCorpora} database have been manually reviewed by the authors on a~case-by-case basis. In~the case of nouns, 26.76\% of all error-producing input words belong to the class of Russian nouns whose animacy cannot be determined outside the context (e.g.~"\textcyrillic{\"еж}" \textemdash  "yozh" \textemdash  "a hedgehog" \textit{ or, depending on context, } "a Czech hedgehog", "\textcyrillic{жучок}" \textemdash  "zhuchok" \textemdash  "a bug" \textit{ or, depending on context,} "a hidden microphone" and the like).
For verbs, 11.26\% of the discrepancies result from the verbs whose perfectiveness cannot be determined outside the context without additional information on the stress in the word (e.g.~"\textcyrillic{насыпать}" \textemdash  "nasypat${}^{'}$" \textemdash  "to~pour~on", "\textcyrillic{пахнуть}" \textemdash  "pakhnut${}^{'}$" \textemdash  "to~smell" \textit{ or, depending on the stress,} "to~smack" etc.).

\par Besides, a~number of errors in \texttt{OpenCorpora} have been discovered.  The classification of flaws in \texttt{OpenCorpora} is beyond the scope of the present work and we only mention that the inflection of the verb
"\textcyrillic{застелить}" \textemdash  "zastelit${}^{'}$" \textemdash  "to~cover" as~well as~the gerund forms of the verbs "\textcyrillic{выместить}" \textemdash  "vymestit${}^{'}$" \textemdash  "to~take revenge~on", "\textcyrillic{напечь}" \textemdash  "napech${}^{'}$" \textemdash  "to~bake", and "\textcyrillic{перекиснуть}" \textemdash  "perekisnut${}^{'}$" \textemdash  "to~go fully sour" appear to be incorrect in this database at the time of writing. In~addition, certain gerunds of a class of reflexive verbs appear to be incorreclty listed in the database.

\par We~remark that the average time needed for the generation of all inflected forms of an adjective is more than ten times shorter than that of a noun despite the fact that the number of forms of an adjective is greater. This fact reflects the high morphological regularity of adjectives in the Russian language whose exceptional inflection is primarily found within a class of possessive  adjectives stemming from animated nouns. 
\par Using the basic functions described above, one can implement automated synthesis of grammatically correct Russian text on the basis of any logical, numerical, financial, factual or any other precise data. The website passare.ru provides examples of such metafunctions that generate grammatically correct weather forecast and currency exchange rates report on the basis of real-time data available online. Besides, it offers a~function that converts a~correct arithmetic formula into Russian~text.
\par Matching adjectives to nouns by gender and number, matching verbs to personal pronouns by person, gender, and number as~well as~numerous similar functions are implemented in the synthesis section of the website. These functions can also be used to put the components of a~sentence into the grammatically correct forms.

\section{Quantitative Corpus Analysis of Russian\\ Morphological Complexity}
\label{sec:corpusAnalysis,Graphs}

\par We now use the algorithms implemented in the web-service www.passare.ru to analyze the complexity of inflection of different parts of speech in the Russian language. There are only three parts of speech that are of interest in this respect, namely, adjectives, nouns, and verbs (together with participles of all kinds). All other parts of speech in the Russian language either comprise a very limited number of words and their forms (like personal and possessive pronouns, conjunctions, interjections etc) or exhibit highly regular inflection (like adverbs). None of these parts of speech are interesting from the algorithmic inflection viewpoint since their irregular inflectional forms are very few and can be easily listed. On~the contrary, inflection of adjectives, nouns and verbs in the Russian language is
highly complex and often irregular (see Fig.~\ref{fig:Reshat-AllForms} for verbs).
\par To measure the morphological variability of a word~$w$ we introduce the function
\begin{equation}
\label{LevenshteinDistance}
\mathcal{L}(w):=\sum\limits_{i,j} {\rm dist}_{L}(w_i,w_j),
\end{equation}
where~$\{w_i\}$ is the list of all forms of the word~$w$ (with a~fixed order of values of grammatical parameters encoding these forms)
and ${\rm dist}_{L}$ is the Levenshtein distance~\cite{Levenshtein} between the forms~$w_i$ and~$w_j.$ For instance, for the verb $w:=$ "\textcyrillic{решать}" \textemdash  "reshat${}^{'}\,$" \textemdash  "to solve" the list~~$\{w_i\}$ of its forms comprises the~78 forms given in Fig.~\ref{fig:Reshat-AllForms}.
\begin{figure*}[htb!]
\centering{}
\includegraphics[width=.8\textwidth]{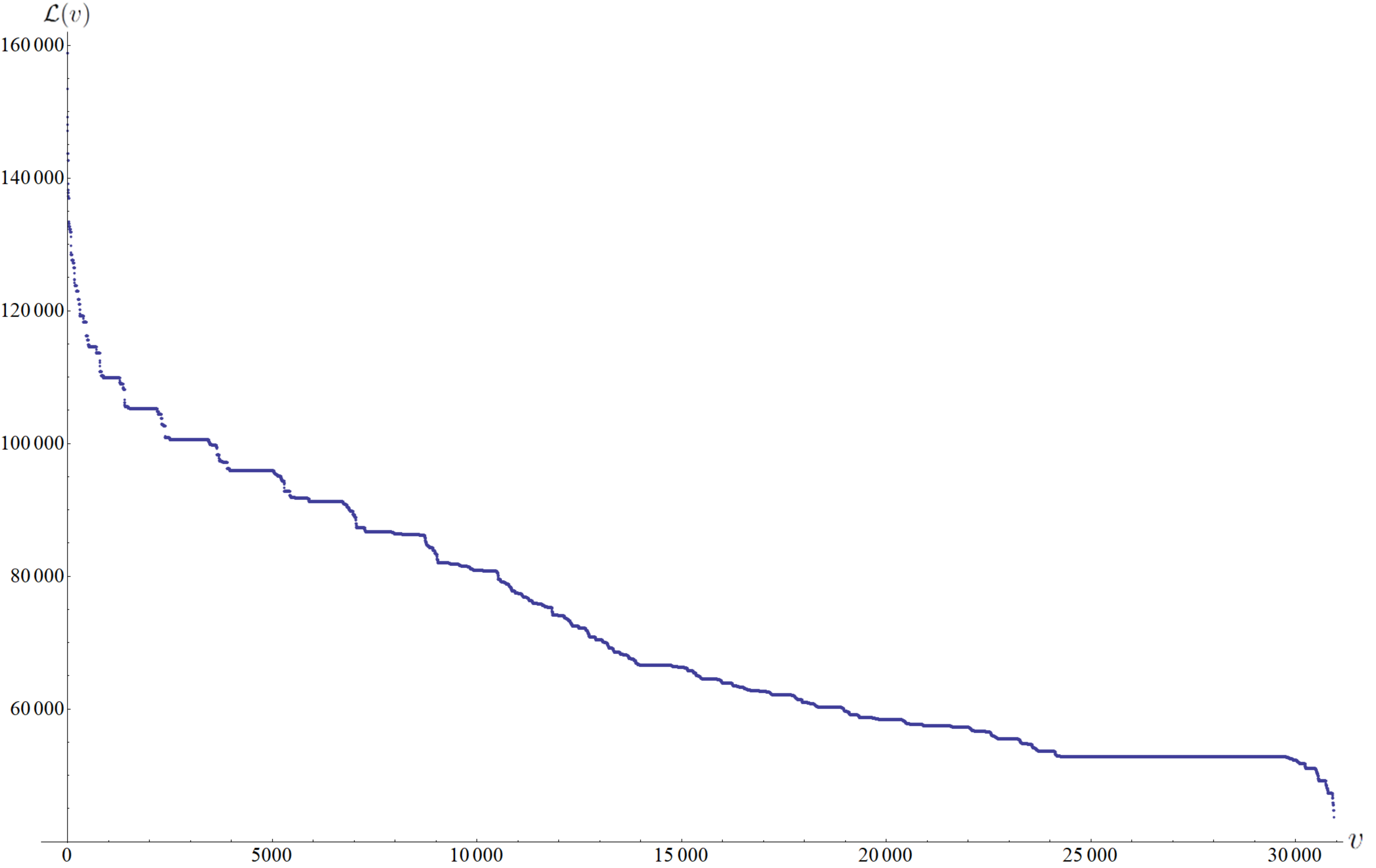}
\caption{Morphological variability of verbs in the Russian language, verbs sorted by the values 
of~$\mathcal{L}(v),$ the total Levenshtein distance~(\ref{LevenshteinDistance}) between the inflected forms of a verb~$v.$ }
\label{fig:verbFormsLevenshteinMonotone}
\end{figure*}

\par \textit{ Verbs.} Verbs exhibit the highest morphological variability among all parts of speech in the Russian
language (cf Fig.~\ref{fig:Reshat-AllForms}). The algorithms for the inflection of verbs and producing various verb forms (participles and gerunds) are among the most complex in Russian grammar.

\par Fig.~\ref{fig:verbFormsLevenshteinMonotone} reflects the morphological variability of verbs in the Russian langauge. The horizontal axes corresponds to the~32358 Russian verbs listed\\ in the \texttt{OpenCorpora} database. The height~$\mathcal{L}(v)$ of a~vertical segment corresponding to a~verb~$v$ has been computed by means of the formula (\ref{LevenshteinDistance}).

\par The forms of a~verb have been computed by means of the inflectional algorithms implemented at www.passare.ru. The performed analysis allows one to detect the Russian verbs (in the \texttt{OpenCor\-po\-ra} database) with the extreme values of their inflectional variability. The overall shape of the monotone curve in Fig.~\ref{fig:verbFormsLevenshteinMonotone} with few flat parts reflects the morphological complexity and very different inflectional patterns of verbs in Russian language. The vast majority of all verbs in the database (more precisely,~69.1\% by our estimates) require detailed case analysis which has been performed in the algorithms implemented in the web-service passare.ru.

\par \textit{ Adjectives.} Adjectives are the part of speech with the most regular inflection in the Russian language. (Here we do not take into account parts of speech with very few words like personal pronouns, interjections, and the like.) Nevertheless, algorithmic inflection of Russian adjectives represents a task of substantial computational complexity.

\par \textit{ Nouns.} In~Russian, nouns exhibit intermediate inflectional complexity compared to adjectives and verbs. Despite the vast majority of regular cases, there exist numerous exceptions which include e.g. indeclinable nouns of foreign origin.

\par A~similar study has been carried out for other parts of speech in the Russian language which has led to a~number of improvements in the inflectional algorithms.

\noindent
\begin{table*}[hb!] 
\centerline{
Table~2: Comparing NLP software that offer Russian inflection or lemmatization}

\small 
\centering
\newcolumntype{L}[1]{>{\raggedright\let\newline\\\arraybackslash\hspace{0pt}}m{#1}}  
	{ 
	\vskip0.2cm \noindent
	\begin{tabular}{|L{2.2cm}|L{2.8cm}|L{2cm}|L{2.2cm}|L{2.8cm}|L{3cm}|}
\hline
	 Software environment & Functionality & Supported languages & Dependency on dictionary  & Distributed as & Implementation \\\hline 
	passare.ru & 
		inflection,
		word matching,  
		data to text
		& Russian & low & free web service & algorithm extraction from language
\\\hline
morpher.ru & inflection (Nouns, Numerals),  simple sentence matching & Russian, Ukrainian & high & commercial web service / standalone libraries & dictionary look-up
\\\hline
	phpmorphy & 
		morphological analysis, 
		lemmatization, 
		inflection 
		& English, Russian, German, Ukrainian, Estonian, and other & high & library (php) & dictionary look-up
\\\hline
	pymorphy2 & morphological analysis,  
    lemmatization, 
		inflection 
		& Russian, Ukrainian & high & library (python) & dictionary look-up
\\\hline
	\end{tabular}
	}
\end{table*}

\noindent
\begin{table*}[ht!] 
\small 
\centering
\newcolumntype{L}[1]{>{\raggedright\let\newline\\\arraybackslash\hspace{0pt}}m{#1}}  
	{ 
	\vskip0.2cm \noindent
	\begin{tabular}{|L{2.2cm}|L{2.8cm}|L{2cm}|L{2.2cm}|L{2.8cm}|L{3cm}|}
\hline
	NooJ & grammar development environment, linguistic analysis 	&	arbitrary & high & framework & grammar based production 
\\\hline 
	MARu & morphological analysis, lemmatization (using pymorphy2) & Russian & high, through pymorphy2 lemmatization & library (python) &
		various machine learning methods: linear model, CRF, deep neural network
\\\hline 
	natasha &
		segmentation, embeddings, morphology, 
		lemmatization, syntax, NER, fact extraction
		& Russian & training data dependency, trained neural models dependency & several libraries (python) & \texttt{razdel} and \texttt{yargy} are rule-based systems; \texttt{navec} and \texttt{slovnet} are neural networks
		\\\hline 
	\end{tabular}
	}
\end{table*}

\section{Sarcasm detection in Russian}

\par The web-service passare.ru offers a sarcasm detection function, aiming at detecting sarcastic sentences in Russian. Automatic detection of sarcasm is a difficult task since it has been a part of any natural language for millennia. It means saying the opposite of what one actually means, sometimes with a distinct tone of voice in a funny way. The suggested process of sarcasm detection is divided into two main steps. 
\par The first step is creating a sentiment analysis model. This model must allow one to classify sentences or text into three classes: positive, neutral, and negative. For this task we use Russian datasets \cite{180}  and machine learning methods \cite{290}. There are three datasets for training and validation. The first one has been collected by Twitter and contains 114991 positive sentences, 111923 negative sentences, and 17639674 untagged sentences. The second one contains  examples of moods from publications in the social network VKontakte manually tagged. The third one has been collected by the authors, using Twitter platform. When sentiment analysis model is ready, it is possible to create a model for sarcasm detection. A neural network model has been built that determines sentiment of sentences divided into 3 classes (negative, neutral, and positive). The accuracy of this model is 79.28\%. \\The confusion matrix is presented in Tab. 3. 
 \newline

 \begin{table}[H] 
 	\begin{center}
Table~3: {Confusion matrix for sentiment analysis}
	\end{center}
\begin{small}
\begin{center}
\begin{tabularx}{0.9\textwidth}{m{1.0cm}XXXX}
\hline
&\tabularxmulticolumncentered{4}{X}{Actual results} \\
\parbox[t]{2mm}{\multirow{4}{*}{\rotatebox[origin=c]{90}{ \parbox{2.9cm}{\centering Predicted \\ results}}}}
&                 &Negative&Positive&Neutral \\
&Negative &1813 &132 &658\\
&Positive &150 &3590 &1089 \\
&Neutral &731 &1072 &9263 \\
\\
\hline
\end{tabularx}
\end{center}
\end{small}
\end{table}
 
\par To build a model for sarcasm detection, there have been collected two datasets of Russian texts, including the hash tag "sarcasm" from Twitter. This dataset contains two classes sarcastic and non sarcastic. It is also divide into two sets. The first set is built to train and test the model. It includes 6598 messages, 5938 are intended for training of the neural network and 660 for validation. 
\par The second set is used to check the results produced by the model. It has not been used in the process of creating sarcasm detection model and contains 3044 sentences for testing. All messages have been preprocessed before training the model. It have been removed special symbols and  words have been converted to lower case and brought to their initial forms using the pymorphy2 library. 
\par There are 42483015 parameters in the architecture of the neural network. This model uses a 500-dimensional embedding to decrease network parameters. There are 49941 examples for training and 5550 examples for validation. The average training time over 3 training epochs on the CPU is 2.1 hours, the total training time on the CPU is 6.3 hours. 
\par During the development process, collected dataset has been divided into two sets. The first set is used for training model and the second one for testing. The training set has been created by the following principle: if a text is classified positively by 3 classes (negative, positive, and neutral) using the tonality dictionary of Russian language \cite{18}, and if there are more positive than negative words in the text, but the text is classified as negative, then it contains sarcasm. The block diagram of the algorithm for sarcasm detection in the sample for training the model is shown in Fig. 5. 
 \begin{figure}[ht!]
 \centering
 \includegraphics[width=0.8\linewidth]{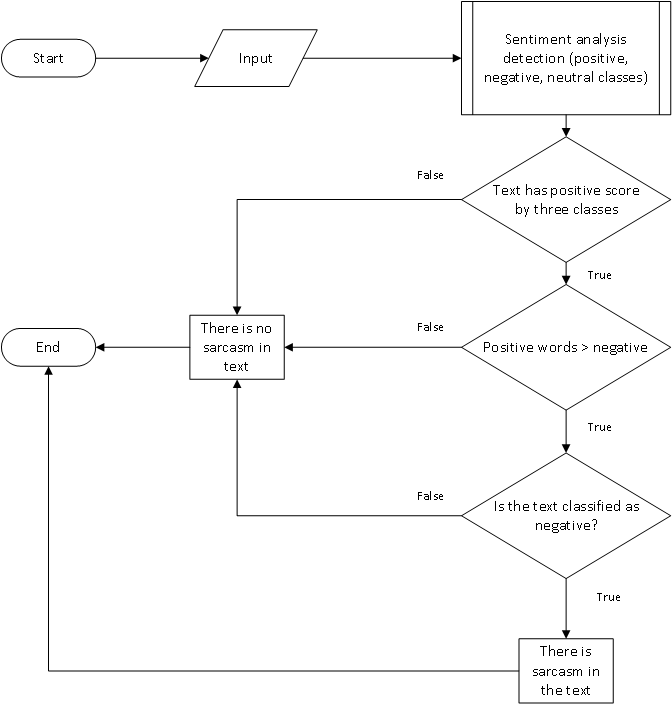}
\caption{The block diagram of the algorithm for determining sarcasm}
 \label{The block diagram of the algorithm for determining sarcasm}
 \end{figure}

\par The model has been trained with the following hyperparameters: burst dimension = 16, maximum message length in tokens = 128, probability of maintaining the weight of connections between layers = 0.5, rate of weight change during training = 1E-05, maximum number of training iterations without reduction = 5. 
\par According to the test results, the final accuracy of the sarcasm model is 78.57\%. After that, the samples for the sarcasm model have been converted taking into account both variants of sarcasm, both negative, expressed in positive words, and positive, expressed in negative ones. After that, the sarcasm model has been retrained on these new samples. The final accuracy on the training set 95.49\%, on the validation set is 90.75\%, and on the test set is 69.19\%. The confusion matrix is presented in Tab. 4.


\begin{table}[H]
	\begin{center}
Table~4: {Confusion matrix for Sarcasm detection model} 
	\end{center}
\begin{small}
\begin{center}
\begin{tabularx}{0.6\textwidth}{m{1.2cm}XXX}
\hline
&\tabularxmulticolumncentered{3}{X}{Actual results} \\
\parbox[t]{2.0mm}{\multirow{4}{*}{\rotatebox[origin=c]{90}{ \parbox{1.5cm}{\centering Predicted \\ results}}}}
&                 &Negative&Positive \\
&Negative &1048 &464 \\
&Positive &474 &1058 \\ 

\\
\hline
\end{tabularx}
\end{center}
\end{small}
\end{table}
 
 \section{Discussion}\label{sec:discussion}

There exist several other approaches towards automated Russian inflection, synthesis of grammatically correct Russian text, and semantic analysis.~\cite{Kanovich-Shalyapina-RUMORS,RussianAndUkrainian}. Besides, numerous programs attempt automated inflection of a~particular part of speech or synthesis of a~document with a~rigid predefined structure~\cite{Chernikov-Karminsky}. Judging by publicly available information, most of such programs make extensive use of manually annotated corpora which might cause failure when the word to be inflected is different enough from the elements in the database. The results of comparison of the approach exposed in the present paper with the other software environments that offer functionality for Russian inflection or lemmatization are summarized in Table~2.

\par The speed of the computationally most expensive inflectional functions of www.passare.ru has been tested against that of the freeware products \texttt{phpmorphy} and \texttt{pymorphy2} on nouns, verbs and adjectives. The corresponding computation times on our system are 3:21, 4:13, and 7:06 for  \texttt{phpmorphy} and 1:00, 2:44, and 1:58 \texttt{pymorphy2} (in min:sec format).

\par The solution presented in this paper has been designed to be as~independent of any dictionary data as~possible. However, due to numerous irregularities in the Russian language, several lists of exceptional linguistic objects (like the list of indeclinable nouns or the list of verbs with strongly irregular gerund forms, see Fig.~\ref{fig:perfectiveGerundFlowchart}) have been composed by the authors and used throughout the code, see {https://github.com/passare-ru/PassareFunctions/} Whenever possible, rational descriptions of exceptional cases have been adopted to keep the numbers of elements in such lists to the minimum.

\section{Acknowledgements}

\par This research was performed in the framework of the state task in the field of scientific activity of the Ministry of Science and Higher Education of the Russian Federation, project "Models, methods, and algorithms of artificial intelligence in the problems of economics for the analysis and style transfer of multidimensional datasets, time series forecasting, and recommendation systems design", grant no. FSSW-2023-0004.

\bibliography{bibliography}
\end{document}